%% file: main.tex
\pdfoutput=1
\documentclass[11pt]{article}

\usepackage{microtype}
\usepackage{graphicx}

\usepackage[T1]{fontenc}
\usepackage[utf8]{inputenc}

\usepackage{booktabs} 
\usepackage{times}
\usepackage{latexsym}
\usepackage{url}
\usepackage{multirow}
\usepackage{balance}
\usepackage{subfigure}
\usepackage{latexsym}
\usepackage{amsmath}
\usepackage{bbm}
\usepackage{paralist}
\usepackage{makecell}

\usepackage{caption}
\usepackage{url}
\usepackage{xspace}
\usepackage{multicol}
\usepackage{multirow}
\usepackage{color}
\usepackage{transparent}
\usepackage{array}
\usepackage{balance}
\usepackage{bm}
\usepackage{tabularx}
\usepackage{hyperref}
\usepackage{graphics}
\usepackage[export]{adjustbox}

\usepackage{emnlp2021}

\definecolor{midnightgreen}{rgb}{0.0, 0.29, 0.33}
\definecolor{darkpink}{rgb}{0.91, 0.33, 0.5}
\definecolor{darkmagenta}{RGB}{139, 0, 139}

\newcommand{\model}{{SEED-Encoder}\xspace}

\usepackage{array}
\newcolumntype{C}[1]{>{\centering\arraybackslash}p{#1}}
\newtheorem{fact}{Fact}

\begin{document}

\title{Less is More: Pre-train a Strong Text Encoder \\
for Dense Retrieval Using a Weak Decoder}

\author{Shuqi Lu$^1$\footnotemark[1] , Di He$^{2}$\footnotemark[2], Chenyan Xiong$^2$\footnotemark[2], Guolin Ke$^2$, Waleed Malik$^2$, Zhicheng Dou$^1$, \\ {\bf Paul Bennett$^2$}, {\bf Tie-Yan Liu$^2$}, {\bf Arnold Overwijk$^2$} \\
        $^1$Renmin University of China $^2$Microsoft \\ 
         \{lusq, dou\}@ruc.edu.cn\\ 
         \{chenyan.xiong, dihe, guolin.ke, waleed.malik,\\ paul.n.bennett, tyliu, arnold.overwijk\}@microsoft.com }







\maketitle

\renewcommand{\thefootnote}{\fnsymbol{footnote}}
\footnotetext[1]{Work done while interning at Microsoft.}
\footnotetext[2]{Corresponding Authors.}

\input{00_abstract}
\input{01_intro}

\input{02_related}

\input{03_method}

\input{04_exp}

\input{05_conclusion}

\input{07_acknowledgements}

\bibliography{emnlp}
\bibliographystyle{acl_natbib}

\newpage
\clearpage
\appendix
\input{06_appendix}

\end{document}

%% file: 00_abstract.tex
\begin{abstract}
Dense retrieval requires high-quality text sequence embeddings to support effective search in the representation space. Autoencoder-based language models are appealing in dense retrieval as they train the encoder to output high-quality embedding that can reconstruct the input texts.
However,  in this paper, we provide theoretical analyses and show empirically that an autoencoder language model with a low reconstruction loss may not provide good sequence representations because the decoder may take shortcuts by exploiting language patterns. To address this, we propose a new self-learning method that pre-trains the autoencoder using a \textit{weak} decoder, with restricted capacity and attention flexibility to push the encoder to provide better text representations. 
Our experiments on web search, news recommendation, and open domain question answering show that our pre-trained model significantly boosts the effectiveness and few-shot ability of dense retrieval models. Our code is available at \url{https://github.com/microsoft/SEED-Encoder/}.

\end{abstract}

%% file: 01_intro.tex
\section{Introduction}

Recently, Dense Retrieval (DR) has progressed to more important roles in many language systems, for example, web search~\cite{xiong2020approximate}, question answering~\cite{karpukhin2020dense},  and news recommendation~\cite{wu2020mind}.
In the first-stage retrieval of these scenarios, DR models generally employ a Siamese/Dual-Encoder architecture in practice. The encoder model first separately encodes the user side (query, browsing history, or question) and the corpus side (document or passages) as individual embeddings in a learned representation space~\citep{lee2019latent}, where retrieval with simple similarity metrics are conducted effectively~\citep{faiss,guo2020accelerating}.

A popular choice of text encoders in DR is the Transformer network pre-trained by language modeling (e.g., BERT)~\citep{reimers-2019-sentence-bert}.
It is unexpected that, unlike in other language tasks where pre-trained models simply excel, directly fine-tuning BERT in DR often underperforms unsupervised sparse retrieval, e.g., BM25. Some complicated procedures are almost necessary to effectively fine-tune pre-trained Transformers in dense retrieval~\citep{karpukhin2020dense, luan2020sparse, xiong2020approximate}. 
One observation is that the pre-trained language models are not effective at encoding the semantics of the entire text sequence in one embedding, especially in dense retrieval where text sequences are mostly longer than 128 tokens~\citep{luan2020sparse}.

In some other modalities, autoencoders have been widely used to obtain high-quality data representations~\citep{vincent2010stacked, kingma2013auto}. They pair a decoder on top of the encoder, trains the decoder to reconstruct the data solely from the encoder's encodings, thus enforce an information bottleneck on the data encodings for better representation quality.
Recently, autoencoders have been brought in language pre-training. \citet{li2020optimus} stacks a GPT-2 decoder on top of the BERT encoder and trains the autoencoder via a conditional language modeling task.
Their learned encoder, Optimus, provides better text encodings for GLUE and language generation tasks, but, as shown in our empirical study, does not provide better encodings for dense retrieval.

This phenomenon inspires us to investigate why the standard setup of autoencoders in language modeling falls short in dense retrieval. 
We first notice that in the auto-regressive decoder, the model takes not only the \texttt{CLS} encoding but also the previous tokens as input. Our mathematical analysis shows that the decoder can exploit natural language patterns using its access to previous tokens and bypass the dependency on the encoder, especially when the sequence is long and the decoder is strong, e.g., GPT-2. As a result, the autoencoder achieving a low reconstruction loss value does not necessarily provide better text sequence encodings.

Our analyses lead to a quite simple solution: we 
 present a new autoencoder pre-training strategy, which pairs the BERT-style encoder with a weak decoder by restricting its parameter capacity and attention flexibility. This way, our \model, ``Strong tExt Encoder by training with weak Decoder'',
 creates an information bottleneck in the autoencoder and forces the encoder to provide better text representations. In our experiments on three real-world applications, we confirm that \model produces better pre-trained checkpoints that seed dense retrieval models with higher accuracy and better few-shot ability.

%% file: 02_related.tex
\section{Related work}
\paragraph{Pre-training Language Models.} Masked Language Modeling (MLM)~\citep{devlin2018bert} is one of the most effective ways to learn text representations. It first randomly masks some tokens in a sequence and then pre-trains a Transformer to recover them~\citep{joshi2020spanbert,liu2019roberta,clark2020electra}. There are also attempts to design sequence-level tasks during pre-training. The next sequence prediction task proposed in \citet{devlin2018bert} trains the model to predict whether two sequences are contiguous. \citet{liu2019roberta} showed this task is not effective and can be removed. In \citet{sun2020ernie}, more sequence-level tasks are developed, such as predicting whether two segments are from the same document. Our learning framework architecture is close to \citet{li2020optimus},  which trains an encoder and a decoder for both language understanding and generation. We will discuss its detail and show how it motivates our work.

\paragraph{Dense Retrieval with Text Encoders.} 
Dense-Retrieval systems often use the Siamese/Dual Encoder architecture, where two sequences are encoded by the Transformer separately, and their similarity is calculated upon their sequence embeddings.  
\citet{reimers2019sentence} is among the first to study how to use BERT in a Siamese architecture and found that the \texttt{CLS} representation does not perform as well as expected. 
Recent research~\citep{karpukhin2020dense, xiong2020approximate} demonstrated that applying pre-trained models in dense text retrieval is not as straightforward.
\citet{karpukhin2020dense} use BM25 to find negative samples to better fine-tune pre-trained models for dense retrieval.
\citet{xiong2020approximate} performs global noise constructive estimation and finds global negatives using the DR model for the DR model.

%% file: 03_method.tex
\section{Method}
In this section, we first recap preliminaries in language pre-training and autoencoder. Then we discuss the drawbacks of using strong decoders in autoencoder and address them with \model.

\subsection{Preliminary}
In a standard setup of pre-training language models, e.g., BERT~\cite{devlin2018bert}, the neural network to be pre-trained is a multi-layer bidirectional Transformer encoder~\citep{vaswani2017attention}, which takes a sequence of tokens $x = (x_1, ..., x_n)$ from the vocabulary $V$, and produces their contextualized representations $\bf{h}=(\bf{h}_1, ..., \bf{h}_n)$:
\begin{align*}
    (\texttt{CLS},x_1, ..., x_n)  \xrightarrow{\text{Transformer}} (\bf{h}_0, \bf{h}_1, ..., \bf{h}_n),
\end{align*}
where \texttt{CLS} is a special token added in the first position, its contextual representation $\bf{h}_0$ is often used as the representation of the sequence. The parameters of the Transformer $\theta_{enc}$ are typically pre-trained using Masked Language Modeling (MLM)~\cite{devlin2018bert}, which masks a fraction of the input sequence and trains the model to predict the original tokens. For ease of reference, we denoted the loss as $\mathcal{L}_\text{MLM}(x,\theta_{enc})$.

As there is no informative training target at the \texttt{CLS} position in token level pre-training tasks, it is not formally guaranteed that the contextual representation at \texttt{CLS} contains enough information for any sequence-level downstream tasks. \citet{li2020optimus} introduces the autoencoder setup in language model pre-training, which adds a reconstruction loss on top of the \texttt{CLS} token's $\bf{h}_0$:
\begin{align}
    x \xrightarrow{\theta_{enc}} \bf{h}_0 \xrightarrow{\theta_{dec}} x.
\end{align}
where $\bf{h}_0$ is viewed as a latent variable. The decoder $\theta_{dec}$, which is another deep Transformer model GPT-2, receives $\bf{h}_0$ and generates the original input autoregressively. The (variational) decoder loss is defined as \cite{li2020optimus}:
\begin{align}
&\mathcal{L}_{dec}(x,\theta_{dec}) = \nonumber \\
&-\sum_{t:1\sim n } \log P(x_t |x_{<t},\bf{h}_0;\theta_{dec}), 
\label{eqn:fulldecode}
\end{align}
where $x_{<t}$ are all previous tokens before $t$.

\subsection{Effects of Using a Strong Decoder}
\label{sec:analysis}
\input{Figures/study_optimus}

One would expect the autoencoder to provide good representations if the decoder can well recover the input. However, we found that a typical model stacking a standard autoregressive decoder on a standard BERT-style encoder doesn't work well in dense retrieval tasks. For example, we fine-tune the pre-trained checkpoint of Optimus, which stacks GPT-2 on top of BERT on MS MARCO and compare it with BERT. 
We use Mean Reciprocal Rank(mrr) and recall as evaluation metrics.
The detailed experimental setting can be found in Section~\ref{sec:web_exp}, and the results are shown in Figure~\ref{fig:optimus_marco}.

The performance of Optimus on dense retrieval tasks is worse than standard BERT, a sharp contrast with Optimus's effectiveness on other language tasks, e.g., in GLUE benchmarks. 
Note that one difference between data in GLUE and MS MARCO is the \emph{sequence length}. In most GLUE tasks, the sequence length is short, e.g., average $14$ tokens in SST-2, while the average passage length in MS MARCO is more than $450$. 
Also, recent research shows that long sentences are hard to represent via single embedding vectors from pre-trained models~\citep{luan2020sparse}. 

To confirm this, 
We randomly select sequence pairs of different lengths and calculate the cosine similarity of their \texttt{CLS} embeddings provided by Optimus. The results are shown in  Figure~\ref{fig:cosine}.
The representations of long sequences (256 or 512 tokens) from Optimus are quite similar; the cosine similarities of random long sequence pairs are around 0.8. 
The model yields cluttered representations for long text sequences. When fine-tuned for dense retrieval in MS MARCO,
 it does not separate relevant documents for a query from those irrelevant ones. All of those representations might be similar to each other and require dedicated fine-tuning to realign their encodings.

\subsection{Theoretical Analysis}

Next, we mathematically show why the encoder may fail to learn good sequence representations using a strong decoder.

In Eqn.~\ref{eqn:fulldecode}, at each time step $t$, the prediction of $x_t$ not only depends on the \texttt{CLS} encoding $\bf{h}_0$  but also the previous tokens $x_{<t}$. Thus a lower reconstruction loss may not be contributed by more informative $\bf{h}_0$: for a large $t$ in a long text sequence, the model may directly predict $x_t$ from $x_{<t}$ if the decoder is strong. The quality of the representation at the \texttt{CLS} is not guaranteed as a low decoding loss may not reflect much about $\bf{h}_0$.

To further understand the requirements for informative sequence representations, we investigate the relationship between the reconstruction loss, $\bf{h}_0$, and the language sequence in their mathematical form. First, we decompose the expectation of the loss $\mathcal{L}_{dec}$ into two terms: a Kullback–Leibler divergence and a conditional-entropy term,  according to the following fact in information theory: 
\begin{fact}
Given two distributions $P(Y,Z)$ and $Q(Y,Z)$ on random variables ($Y$, $Z$), we have 
\begin{align}
    \operatorname{E}&_{Y,Z\sim P}[-\log Q(Z|Y)] \nonumber \\
    =&\operatorname{E}_{ Y\sim P(Y)}[D_\text{KL}(P(Z|Y) ||Q(Z|Y))] 
    \\ &+H_P(Z|Y).\nonumber
\end{align} 
\end{fact}
We have $X$ as a random variable defined in the sequence space $\mathcal{X}$,  where each sequence $x$ is sampled from data distribution $P_D$, $X_{<t}$ as the truncate of $X$ at position $t$, and $P_{\theta_{dec}}$ as the sequence distribution generated by the decoder. For simplicity, we assume all the sequences are of length $n$. The expected reconstruction loss can be rewritten as 
\begin{align}
\operatorname{E}&_{D}[\mathcal{L}_{dec}(X,\theta_{dec})]  \label{eqn:needweak}\\
=& \operatorname{E}_{D}\left[ \sum_{t:1 \sim n}  -\log P(X_t|X_{<t},\bf{h}_0;\theta_{dec}) \right]\\
=&\sum_{t:1 \sim n} \operatorname{E}_{D} \Big[ D_\text{KL}\Big(P_D(X_{t}|X_{<t},\bf{h}_0)|| \label{eq:kl1} \\
& \qquad \quad   P_{\theta_{dec}}(X_{t}|X_{<t},\bf{h}_0)\Big)\Big]  \label{eq:kl2} \\
&  \qquad \quad  + H_{D}(X_t|X_{<t}, \bf{h}_0). \label{eq:h} 
\end{align}

The above equation shows that the loss consists of two terms, a K-L term $D_\text{KL}(\cdot)$ (Eqn.~\ref{eq:kl1} and Eqn.~\ref{eq:kl2}) describing the difference between two distributions, and a conditional-entropy term $H_D(\cdot)$ (Eqn.~\ref{eq:h}) reflecting the strength of language patterns. As we discuss next, both terms can achieve low values even with random $\bf{h}_0$. 

The first K-L term characterizes how $P_{\theta_{dec}}(X_{t}|X_{<t},\bf{h}_0)$,  the decoder generated sequence distribution, aligns with the ground truth distribution $P_D(X_{t}|X_{<t},\bf{h}_0)$. 
Even with a meaningless $\theta_{enc}$, if the decoder has sufficient capacity, e.g., a very deep Transformer, it can still approximate the ground truth distribution well and thereby reduce the K-L term. In theory, Transformers with arbitrary width and depth can approximate any sequence-level functions and may reach a low K-L loss using little information from $\bf{h}_0$~\cite{yun2019transformers}.

The second term $H_{D}(X_t|X_{<t}, \bf{h}_0)$ characterizes the strength of language patterns: The stronger the correlation between $X_t$ with $X_{<t}$, the lower the second term is. In natural language, the correlation becomes stronger with larger $t$ as there is more information from the previous tokens. There is not a strong need for a good text encoder  $\bf{h}_0$ because a strong decoder can capture the natural language patterns by itself.

\input{Figures/model}
\subsection{SEED-Encoder}
\label{sec:method}
Our analysis shows that to obtain a stronger text encoder and a better $\bf{h}_0$, we can not make the decoder too strong: we need to constrain its capacity and also the available language context to reduce the correlation between $X_t$ and $X_{<t}$, so that it has to rely on the information in the encoder \texttt{CLS} to reconstruct the text sequence.

In the rest of this section, We introduce \model which adopts these designs. The model structure is illustrated in Figure~\ref{fig:model}. 

Making a language model weaker is easier than making it stronger. We simply modify Eqn.~\ref{eqn:fulldecode} to weaken the decoder: 
\begin{itemize}
    \item  Using a shallower Transformer $\theta^{weak}_{dec}$ with fewer layers (e.g., three);
    \item  Restricting its access to previous context, i.e., limit model attention to previous $k$ tokens. 
\end{itemize}
This leads to the following reconstruction loss:
\begin{align}
&\mathcal{L}^{weak}_{dec}(x,\theta^{weak}_{dec}) = \nonumber \\
&-\sum_{t:1\sim n } \log P(x_t|x_{t-k:t-1},\bf{h}_0;\theta^{weak}_{dec}),
\end{align}
where $k$ is the window size of the restricted attention.
Through these modifications, we enforce the information bottleneck between the encoder and the decoder, thereby forcing the decoder to rely on the \texttt{CLS} representation of the encoder, and pushing the encoder to learn a more informative representation.

Similar to \citet{li2020optimus}, the pre-training of \model uses the combination of the encoder's standard MLM loss and the decoder's reconstruction loss:
\begin{align}
 &\mathcal{L}(x,\theta_{enc}, \theta^{weak}_{dec}) =  \nonumber \\
& \mathcal{L}_\text{MLM}(x,\theta_{enc}) + \mathcal{L}^{weak}_{dec}(x,\theta^{weak}_{dec}).
\end{align}
The encoder and decoder are trained together.  After pre-training, the decoder is discarded, and the encoder is used in downstream applications.

%% file: Figures/study_optimus.tex
\begin{figure}[t]
\centering
\vspace{-0.1cm}
\subfigure[Ranking accuracy\label{fig:optimus_marco}]{
\begin{minipage}[t]{0.465\linewidth}
\centering
\includegraphics[width=3.8cm]{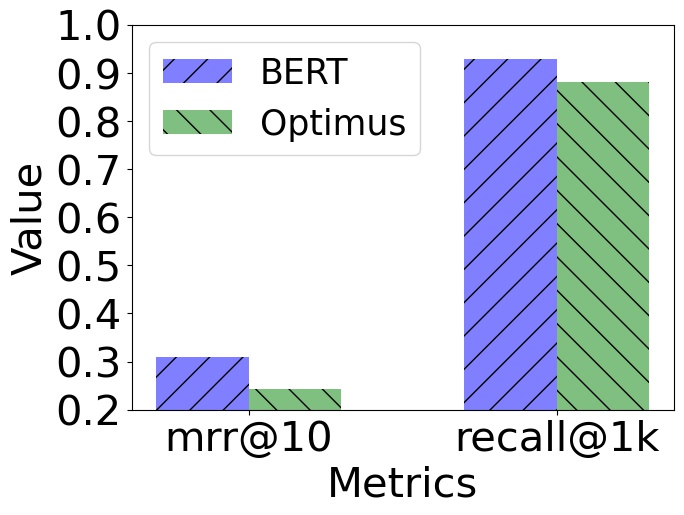}
\end{minipage}
}
\subfigure[Cosine similarity\label{fig:cosine}]{
\begin{minipage}[t]{0.465\linewidth}
\centering
\includegraphics[width=3.8cm]{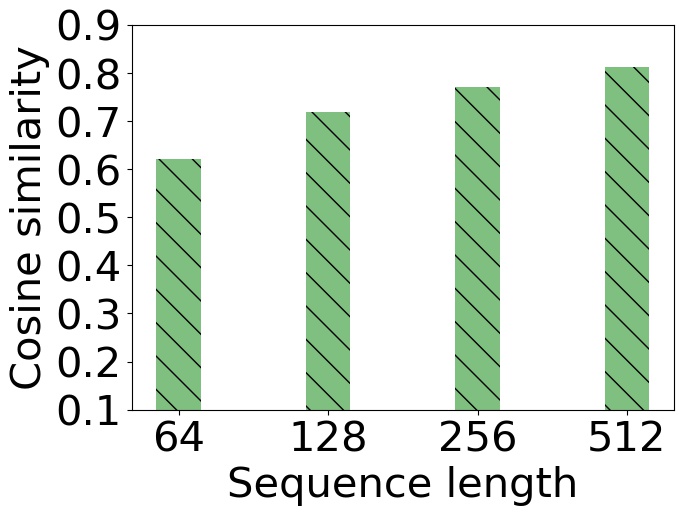} 
\end{minipage}
}
\caption{Behaviors of Optimus on MS MARCO Passage Ranking Dev set: (a) its ranking accuracy in comparison with vanilla BERT; (b) its sequence representations' cosine similarity at variant lengths. \label{fig:study_optimus}}

\vspace{-0.4cm}
\end{figure}

%% file: Figures/model.tex
 \begin{figure}[!tbp]%
\includegraphics[width=1.0\linewidth]{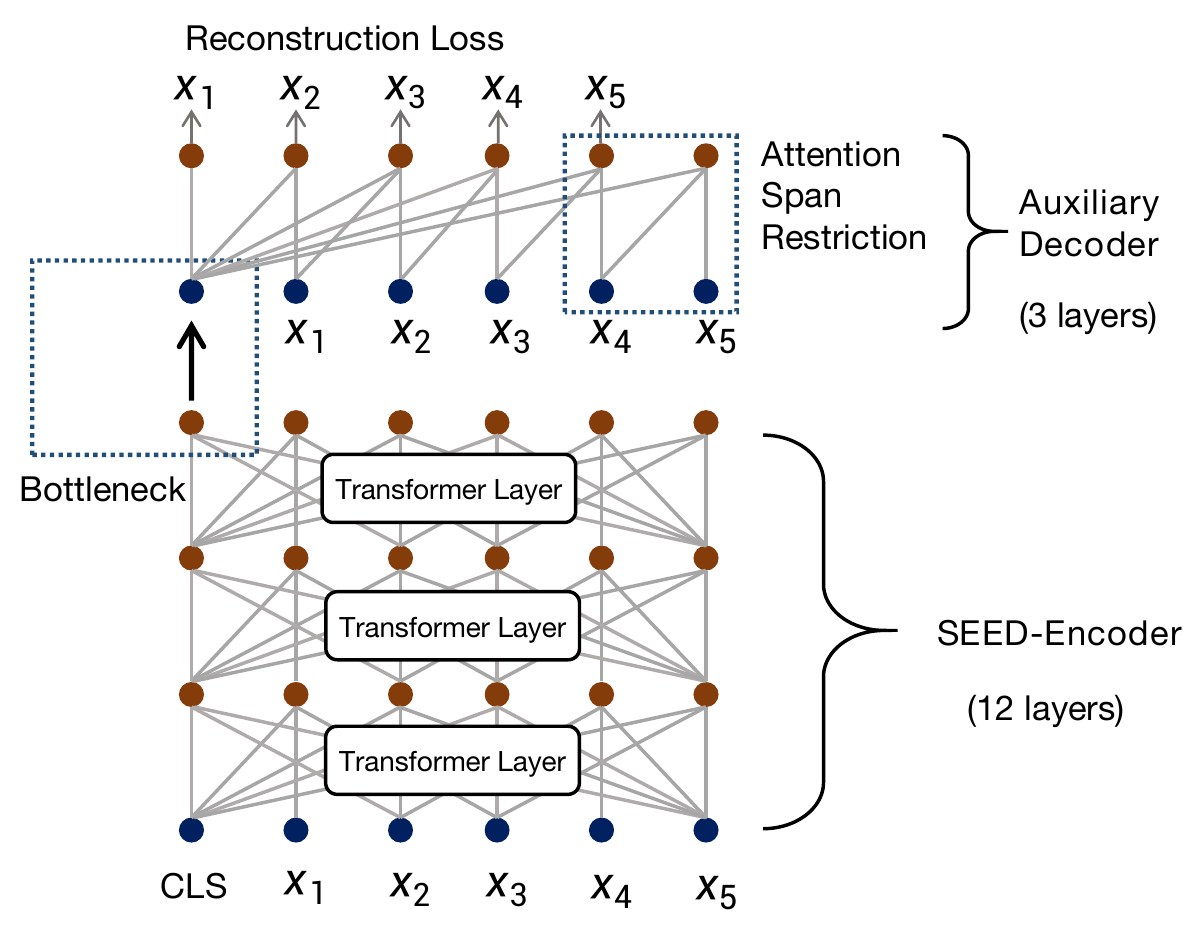}%
\caption{The structure of \model with an auxiliary decoder. The encoder and decoder are connected only via the [CLS] representation as the information bottleneck. The decoder capacity is restricted in both parameter size and attention span.\label{fig:model}}

\end{figure}%

%% file: 04_exp.tex
\section{Experiments}
In this section, we present various experimental analyses to evaluate the  \model on dense retrieval tasks.  More results on other language tasks are in Appendix~\ref{sec:glue}. 
\input{tables/passage_result}

\subsection{Pre-training Details}
\label{sec:exprimentdesign}

All our models are pre-trained \textit{from scratch}, following the setup of BERT-base~\citep{devlin2018bert}: pre-training on English Wikipedia and BookCorpus \citep{moviebook} (roughly 16GB texts) for 1M steps, with batch size 256, maximum sequence length 512, and 15\% masks. We follow the pre-processing steps and use 32,768 sub-word tokens in \citet{ke2020rethinking}. We remove the next sentence prediction task following \citet{liu2019roberta}.  

We use Adam \citep{DBLP:journals/corr/KingmaB14} as the optimizer, and set its hyperparameter $\epsilon$ to 1e-6 and $(\beta1, \beta2)$ to (0.9, 0.999). The peak learning rate is set to 1e-4 with a 10$k$-step warm-up stage. After the warm-up stage, the learning rate decays linearly to zero. We set the dropout probability to 0.1, gradient clip norm to 1.0, and weight decay to 0.01. All codes are implemented based on \emph{fairseq} \citep{ott2019fairseq} in \emph{PyTorch} \citep{paszke2017automatic}. All models are run on 8 NVIDIA Tesla V100 GPUs with mixed-precision \citep{micikevicius2017mixed}. 

Our encoder architecture is the same with BERT-base: 12 Transformer layers, eight attention heads, and 768 hidden dimensions (110M parameters). We use a three-layer Transformer as the decoder, restrict its attention to the previous two tokens (attention span $k=2$), and keep all else the same with the encoder. 
The decoder is only used in pre-training and is dropped during fine-tuning. There is no additional cost in fine-tuning nor inference. 



\subsection{Fine-tuning Siamese/Dual-Encoders}
\label{app:finetune}
Fine-tuning \model in the Siamese architecture on the dense retrieval tasks is the same as other pre-trained models.  Here we show how fine-tuning in a typical sentence pair matching task with binary labels can be done with Triplet loss.
\begin{align*}
    \mathcal{L} \!  = \!\!  \sum_{x^q, x^{d+}, x^{d-}} \!\!\!\!\! \text{relu}(1 - (s(x^q, x^{d+}) - s(x^q, x^{d-}))).
\end{align*}
The training data include triples of query $x^q$ and its positive/negative labeled sequence $(x^{d+}, x^{d-})$.
The scoring of the sequence pairs $s(x^q, x^{d})$ is done by simple similarity functions, such as cosine and dot product, on their \texttt{CLS} encodings. 
More advanced fine-tuning strategies~\citep{karpukhin2020dense, xiong2020approximate} can also be used as \model is an alternative for other pre-trained encoders.

\subsection{Experiments on Web Search}
\label{sec:web_exp}
Our first application,  web search~\citep{lee2019latent},.uses the MS MARCO~\citep{bajaj2016ms} dataset, the largest public search benchmark to date.
It includes two tasks, passage ranking and document ranking. We focus on the first stage retrieval step, which is to find relevant passages/documents from the entire corpus. We also show the results in the reranking setting where all models rerank a pre-given set of candidate documents (Top 100 from BM25) for reference. More details of MARCO are in Appendix~\ref{app:dataset}.

Our pre-trained encoders are fine-tuned with ANCE negative sampling strategy~\citep{xiong2020approximate}. In document retrieval, we use ANCE (FirstP) which uses the first 512 tokens of the long document and cut-off the rest.
We also evaluate with another negative sampling strategy, BM25 Neg, which uses top 100 BM25 retrieved results as negatives samples and performs similar to DPR~\citep{karpukhin2020dense} on MARCO.


\paragraph{Baselines:} The main baseline is our run of BERT-base~\citep{devlin2018bert,liu2019roberta}, which we pre-trained and fine-tuned in the exact setting with \model. We use the permutation test and $p<0.05$ as the statistical significance test between \model and BERT (Ours). Besides BERT, we evaluate two other pre-trained language models in the same setting: ELECTRA~\cite{clark2020electra} and ERNIE2.0~\cite{sun2020ernie}. ELECTRA is one of the most effective pre-trained encoders on the GLUE benchmark~\citep{clark2019electra}. ERNIE2.0 uses various token-level tasks and sentence-level tasks, including an IR Relevance Task. We use the MARCO passage benchmark to showcase the performance of these two pre-trained models.

In addition, we also list the task-specific first stage retrieval baselines that were published recently or submitted to the leaderboard, although they barely outperform our vanilla BERT baseline. 
For passage ranking, the classic sparse retrieval baselines include the standard BM25, Best TREC Sparse Retrieval with tuned query expansion, and Best DeepCT, all from TREC DL 2019 official evaluation~\cite{craswell2020overview}. 
These three approaches represent the standard sparse retrieval, best classical sparse retrieval, and the latest method of using BERT to improve sparse retrieval. 

For document ranking, BM25~\cite{craswell2020overview} and the enriched traditional IR baseline are standard sparse retrieval baselines. The enriched traditional IR baseline uses pre-defined IR features, including BM25, to rank the documents.
BM25 + doc2query-T5 expansion uses Doc2query model\cite{nogueira2019document}, expanding the documents with predicted queries that are related to or representative of the documents' content. The queries are predicted by a sequence-to-sequence model taking the document terms as input. 
Both DE-hybrid and ME-hybrid~\cite{luan2020sparse} use dense features from BERT and hand-craft sparse features. DE-hybrid takes the \textit{CLS} representations of document and query as the dense feature and calculates the dot product similarity. This similarity score is further combined with sparse retrieval scores as the final score for ranking. Different from DE-hybrid, ME-hybrid uses max-pooling over multiple contextual embeddings as dense features.

\input{tables/doc_result}

\paragraph{Results:} The results of \model and baselines in MARCO Passage retrieval and Doc retrieval are listed in Table~\ref{tab:passage-res} and Table~\ref{tab:doc-res}. \model outperforms all existing baselines on all benchmarks. By simply switching its fine-tuning starting checkpoint from BERT to \model---without changing any architectures nor fine-tuning strategies---the accuracy is significantly improved on these two large-scale benchmarks. 

In comparison, on MARCO Passage retrieval, switching from BERT to ELECTRA or ERNIE2.0 does not improve the retrieval accuracy. Pre-training models optimized for other scenarios are not necessarily better for dense retrieval.

On MARCO document retrieval,
ANCE (FirstP) only uses one vector per document from its first passage, while ANCE (MaxP) uses four vectors per document from its first four passages, which often cover the full document body.
Yet with \model as the starting point, ANCE (FirstP)  outperforms the recent state-of-the-art ANCE (MaxP) with RoBERTa by relatively 6\% in the hidden Eval, while using fewer embeddings per document.
Reducing embeddings required per document is important in real search systems where the corpus size is beyond billions~\citep{xiong2020approximate}.

\input{tables/mind_result}

\subsection{Experiments on News Recommendation}

Our second application is news article recommendation, another important real-world task that connects users with information. We use the recently released MIcrosoft News Dataset (MIND) benchmark~\cite{wu2020mind}. The task is to rank a given set of candidate news articles based on the user's previous click history on MSN news articles. The evaluation uses the user's click as the positive label. We use the public MIND Dev and its official metrics: AUC, MRR, NDCG@5, and NDCG@10. More details of MIND are in Appendix~\ref{app:dataset}.

We follow MIND's official setting and use a standard dense retrieval model to rerank the pre-given candidate news articles. 
Our DR model represents each user's history by concatenating all the titles they clicked on the MSN site, with [SEP] tokens in between, and using as many recent titles as possible within the 512 length limit.
The candidate articles are represented by the concatenation of their titles and snippets.
Then it  encodes the user history and candidate articles with \model, and matches them with dot-products. 

\paragraph{Baselines:} MIND is a relatively new benchmark. The most recent baselines are those in \citet{wu2020transformer}. Based on Transformer~\cite{vaswani2017attention}, Transformer-XL~\cite{dai2019transformer} uses relative positional encodings that
integrate content-dependent positional scores and
a global positional score in the self-attention layer.
TENER~\cite{yan2019tener} uses direction-aware sinusoidal relative position embeddings in a similar way as in Transformer-XL.
Different from Transformer-XL and TENER, DA-Transformer~\cite{wu2020transformer} directly re-scales the attention weights based on the mapped relative distances instead of using sinusoidal position embeddings. 
Similar to the web search experiments, we also compare \model with BERT (Ours).

\paragraph{Results:} The results of \model and baselines in  MIND are listed in Table~\ref{tab:mind-res}. \model outperforms the recent state-of-the-art DA-Transformer, which employs various architecture improvements specifically designed for recommendation~\citep{wu2020transformer}. A better self-learning strategy to leverage unsupervised data can be as effective as, if not better than, task-specific architecture changes while avoiding all the engineering hustles.

\input{tables/nq_results}
\subsection{Experiments on Open QA}
Our third application is dense retrieval in open-domain question answering. This task often leverages a two-stage framework:  first uses a context retriever
to select a small set of passages that may contain the answer to the question; and
then uses a machine reader which thoroughly examines the retrieved passages and identifies the correct
answer ~\cite{karpukhin2020dense}. We focus on the first stage, i.e., dense retrieval to select relevant passages. We use Natural Question query set~\cite{kwiatkowski2019natural} and the Wikipedia passages prepared and shared in DPR~\cite{karpukhin2020dense}. More details of the NQ dataset are in Appendix~\ref{app:dataset}.
We follow the evaluation metrics used in DPR, hit accuracy of Top-20 and Top-100. 

Models are fine-tuned using DPR fine-tuning strategy as in ~\citet{karpukhin2020dense}, which uses a dual-encoder architecture and samples negatives in the mini-batch. We also experiment with the ANCE fine-tuning strategy as in ~\citet{xiong2020approximate} which dynamically samples hard negatives.

\paragraph{Baselines: }
We take BM25, BERT as baselines as in~\citet{karpukhin2020dense}.  Consistent with the web search tasks and news recommendation tasks, we also compare \model with BERT (ours).

\paragraph{Results: }
The results of \model and the baselines on NQ benchmark are in Table~\ref{tab:nq-res}. Again, \model outperforms all other baselines with DPR negative sampling or ANCE negative sampling. We do not change any architectures nor fine-tune strategies and only simply switch the BERT checkpoint to \model, but bring significant improvements on the large-scale benchmark.

\subsection{Discussion and Analysis}
In this section, we conduct more analysis to understand the advantages of the \model. For simplicity, all experiments are run on the MS MARCO document retrieval tasks.
\input{Figures/ablation}

\subsubsection{Ablation study}
In the experiments above, we use a three-layer Transformer decoder and restrict the attention span to be two. One may wonder whether such constraints are essential for learning good sentence representations. In this section, we try various decoder configurations with different numbers of layers and attention window sizes. 


From the results in Figure~\ref{fig:ablation}, we can see that the \model with the stronger decoder, 5-layer Transformer with full attention (All), performs worse than those with weaker decoders in dense retrieval. The retrieval accuracy correlated well with the decoder capacity of the corresponding \model.
So unlike typical multi-task settings where tasks share lower-level representations and correlate in accuracy, in \model, the decoder is to force the encoder to capture more information in its sequence embeddings: A weak decoder leads to a stronger encoder.

\input{Figures/dependency}

To further understand the relationship of  encoder's \texttt{CLS} embedding and the decoder, in Figure~\ref{fig:whole_dependency} we plot the cosine similarity between the decoder's token representations in its last layer and the encoder's \texttt{CLS}. 
The impact of restricting attention is significant: with full attention (Figure~\ref{fig:dependency}), the decoder may depend heavily on the encoder's \texttt{CLS} in the beginning but quickly drops the dependency when sufficient context information is available; with restricted access to context, the decoder is forced to attend more on the encoder's \texttt{CLS} representation in all token positions, as shown in the consistent cosine similarity in different positions in Figure~\ref{fig:dependency_window}. This confirms that when the decoder is weak (restricted attention), it depends more on the encoder's \texttt{CLS}, thus pushes the encoder to learn more informative representations. Also, the results in Figure~\ref{fig:dependency} suggest that when using a powerful encoder, the CLS embedding will encode the first several words in the sentence but might ignore others. This can be one of the reasons that Optimus performs worse than BERT in dense retrieval in Figure~\ref{fig:optimus_marco}.

\input{Figures/compare_optimus}

\input{Figures/startingpoint}
\subsubsection{Document Representation Quality }
    

In Section 3.2, we empirically show that using a standard autoencoder learning framework, the similarity between sequence representations grows to be large for long sequences. In this section, we first study whether \model improves the representation diversity. Similar to Figure~\ref{fig:cosine}, we collect randomly sampled sentence pairs and calculate the cosine similarity of their \texttt{CLS} encodings generated by \model.

Results in Figure~\ref{fig:compare_optimus} shows that, the \texttt{CLS} embedding generated by \model is more diverse. The average \texttt{CLS} cosine similarity is only $0.48$ even when the sentence length is $512$. This result shows that  \model can well differentiate sentences during pre-training.

\paragraph{Few-shot effectiveness} 
Note that diverse representations don't necessarily mean high-quality. To figure out the effectiveness of the representation, we conduct few-shot learning experiments for \model. In particular, we record the dev performance during the fine-tuning stage and check how many training iterations and how many samples are required for the model to achieve a reasonably good performance. 

In Figure~\ref{fig:s_bm25} and~\ref{fig:s_ance}, 
we plot the retrieval accuracy at different fine-tuning steps. Starting from \model instead of BERT, both the vanilla Siamese and ANCE achieve higher retrieval accuracy in the very beginning and maintain their advantages throughout the fine-tuning process. 
For example, Siamese (BM25 Neg) only requires 30k fine-tuning iterations with \model to reach BERT's best performance at 140k iterations. With ANCE (First P), it takes 150K iterations with \model versus 750K with BERT. 

In Figure~\ref{fig:fewshot_bm25} and ~\ref{fig:fewshot_ance}, we plot the retrieval accuracy with different fractions of training data. Compared with BERT, with fewer training labels, \model always reaches better accuracy. When only using $10\%$ training labels, \model (MRR 0.318 in Figure~\ref{fig:fewshot_bm25}) is still competitive with BERT using all training labels (MRR $0.32$).

These results indicate that the representation learned by \model is better than that learned by BERT.  The reduction in fine-tuning cost helps democratize the benefits of pre-training models, especially in applications where computing resources or task-specific supervision is restricted. 

\input{tables/case_study}

\paragraph{Case Study} 
We further showcase some winning examples of \model in Table~\ref{tab:case}.
The error made by BERT correlated with our observation in Figure~\ref{fig:dependency}, where the encoder's representation is more related to those tokens at the beginning of the text sequences, which is quite related to the query.
Only when the model captures the information of the entire text can it find the correct documents. For example, in the first case, \model captures ``winter hiking'' at the back of the document while BERT only pays attention to some of the keywords at the beginning of the document even if the overall semantics does not match, and in the second case, BERT missed the "party" part in the query.

%% file: tables/passage_result.tex
\begin{table}[t]
\vspace{0.18cm}
\small
\centering
\resizebox{0.48\textwidth}{!}{
\begin{tabular}{l|r|rr}
\hline
    {} & \textbf{Rerank} & \multicolumn{2}{c}{\textbf{Retrieval}}\\
    \hline
    \textbf{Model} &
    \multicolumn{1}{c|}{\textbf{MRR@10}} & \multicolumn{1}{c}{\textbf{MRR@10}} & \multicolumn{1}{c}{\textbf{Recall@1k}}  \\
    \hline
    BM25~\cite{craswell2020overview} & - & 0.240 & 0.814\\
    Best DeepCT~\cite{dai2019context} &- & 0.243 & n.a. \\
    Best TREC Trad IR~\cite{craswell2020overview} & - & 0.240 & n.a.\\
    DPR (RoBERTa)~\cite{karpukhin2020dense} & - & 0.311 & 0.952\\
    \hline
    \multicolumn{4}{l}{\textbf{With DPR (BM25 Neg)}}\\
    \hline
    BERT~\cite{devlin2018bert} & 0.317 & 0.310 & 0.929 \\
    Optimus~\cite{li2020optimus} & 0.300 & 0.244 & 0.880 \\
    ELECTRA~\cite{clark2020electra} & 0.300 & 0.258 & 0.854 \\
    ERNIE2.0~\cite{sun2020ernie} & 0.324 & 0.321 & 0.942 \\
    RoBERTa~\cite{liu2019roberta} &- & 0.299 & 0.928  \\
    BERT (Ours) & 0.326 & 0.320 & 0.933  \\
    SEED-Encoder & $\textbf{0.329}^{\dagger}$ & $\textbf{0.329}^{\dagger}$ & $\textbf{0.953}^{\dagger}$  \\
    \hline
    \multicolumn{3}{l}{\textbf{With ANCE (FirstP)}}\\
    \hline
    RoBERTa~\cite{liu2019roberta} &- & 0.330 & 0.959 \\
    BERT (Ours) & 0.327 & 0.332 & 0.952  \\
    SEED-Encoder & $\textbf{0.334}^{\dagger}$ & $\textbf{0.339}^{\dagger}$ & $\textbf{0.961}^{\dagger}$  \\
    \hline
\end{tabular}
}
\caption{First stage retrieval results on MS MARCO Passage ranking Dev set. Rerank MRR is for reference only. 
Statistically significant improvements over BERT (Ours) are marked by $\dagger$. 
}
\label{tab:passage-res}
\vspace{-0.3cm}
\end{table}

%% file: tables/doc_result.tex
\begin{table}[t]
\small
\resizebox{0.48\textwidth}{!}{
\begin{tabular}{l|rr|r}
\hline
  	  & \multicolumn{2}{c|}{\textbf{Dev}} & \multicolumn{1}{|c}{\textbf{Eval }}   \\
   \hline
    \textbf{Model} &
     \textbf{Rerank} & \textbf{Retrieval} & \textbf{Retrieval}\\
    \hline
    BM25~\cite{craswell2020overview} & - & 0.318 & 0.284\\
    DE-hybrid~\cite{luan2020sparse} & - & - & 0.287\\
    BM25 + doc2query-T5 expansion & - & 0.327 & 0.291\\
        ME-hybrid~\cite{luan2020sparse} & - & - & 0.310 \\
        Enriched Traditional IR Baseline & - & 0.355 & 0.312\\
    ANCE MaxP (RoBERTa)~\cite{xiong2020approximate} & - & 0.384 & 0.342\\
    \hline
    \multicolumn{4}{l}{\textbf{With DPR (BM25 Neg)}}\\
    \hline
    BERT (Ours) & 0.338 & 0.308 & -\\
    SEED-Encoder&  $\textbf{0.344}^{\dagger}$ & $\textbf{0.323}^{\dagger}$ & - \\
    \hline
    \multicolumn{4}{l}{\textbf{With ANCE (FirstP)}}\\
    \hline
    RoBERTa~\cite{liu2019roberta} & -  & 0.373 & - \\
    BERT (Ours) & 0.368 & 0.382 & -\\
    SEED-Encoder & $\textbf{0.377}^{\dagger }$ & $\textbf{0.394}^\dagger $ & \textbf{0.362} \\
    \hline

\end{tabular}
}
\caption{MRR@100 on MARCO Documents from first-stage retrieval methods. Rerank results are for reference only. 
Statistically significant improvements over BERT (Ours) are marked by $\dagger$. 
}
\label{tab:doc-res}
\vspace{-0.3cm}
\end{table}

%% file: tables/mind_result.tex
\begin{table}[t]
\small
\centering
\resizebox{0.48\textwidth}{!}{
\begin{tabular}{p{0.57\columnwidth}|p{0.09\columnwidth}p{0.09\columnwidth}p{0.16\columnwidth}<{\centering} p{0.16\columnwidth}<{\centering}}

\hline
    \textbf{Model} & \textbf{AUC} & \textbf{MRR} & \textbf{NDCG@5} & \textbf{NDCG@10}\\
    \hline
    Transformer~\cite{vaswani2017attention} & 0.6776 & 0.3305 & 0.3594 & 0.4163\\
    Transformer-XL~\cite{dai2019transformer} & 0.6792 & 0.3315 & 0.3604 & 0.4170\\
    TENER~\cite{yan2019tener} & 0.6770 & 0.3301 & 0.3589 & 0.4158\\
    DA-Transformer~\cite{wu2020transformer} & 0.6832 & 0.3336 & 0.3634 & 0.4207\\
    \hline
     \multicolumn{4}{l}{\textbf{With DPR (MIND Neg)}}\\ \hline
    BERT (ours) & 0.7015 & 0.346  & 0.3844 & 0.4479\\
    SEED-Encoder & $\textbf{0.7059}^{\dagger}$ & $\textbf{0.3506}^{\dagger}$ & $\textbf{0.3908}^{\dagger}$ & $\textbf{0.4526}^{\dagger}$ \\
    \hline

 \end{tabular}
 }
\caption{Results on MIND news recommendation benchmark. All methods are evaluated in the reranking setting with pre-given news candidates in MIND, to follow their official setting. Baseline scores are obtained from \citet{wu2020transformer}. Statistically significant improvements over BERT (Ours) are marked by $\dagger$.
}
\label{tab:mind-res}
\vspace{-0.3cm}
\end{table}

%% file: tables/nq_results.tex
\begin{table}[t]
\small
\resizebox{0.48\textwidth}{!}{
\begin{tabular}{l|rr}
\hline
    \textbf{Model} &
     \textbf{Top-20} & \textbf{Top-100} \\
    \hline
    BM25~\cite{craswell2020overview} & 59.1 & 73.7 \\
    \hline
    \multicolumn{3}{l}{\textbf{With DPR}}\\
    \hline
    BERT~\cite{karpukhin2020dense} & 78.4 & 85.4 \\
    BERT (BM25 +DPR)~\cite{karpukhin2020dense} & 76.6 & 83.8 \\
    BERT  (Ours) & 77.8 & 85.1 \\
    SEED-Encoder &  $\textbf{80.4}^{\dagger}$ & $\textbf{87.1}^{\dagger}$ \\
    \hline
    \multicolumn{3}{l}{\textbf{With ANCE}}\\
    \hline
    BERT~\cite{xiong2020approximate} & 81.9 & 87.5\\
    SEED-Encoder & $\textbf{83.1}^{\dagger }$ & $\textbf{88.7}^\dagger $\\
    \hline
\end{tabular}
}
\caption{Retrieval results (Answer Coverage at Top-20/100) on Natural Questions in the
setting from~\cite{karpukhin2020dense}.
Statistically significant improvements over BERT are marked by $\dagger$.  
}
\label{tab:nq-res}
\vspace{-0.3cm}
\end{table}

%% file: Figures/ablation.tex

\begin{figure}[t]
\centering
\vspace{-0.22cm}
\subfigure[MRR@10\label{fig:mrr}]{
\begin{minipage}[t]{0.465\linewidth}
\centering
\includegraphics[width=3.8cm]{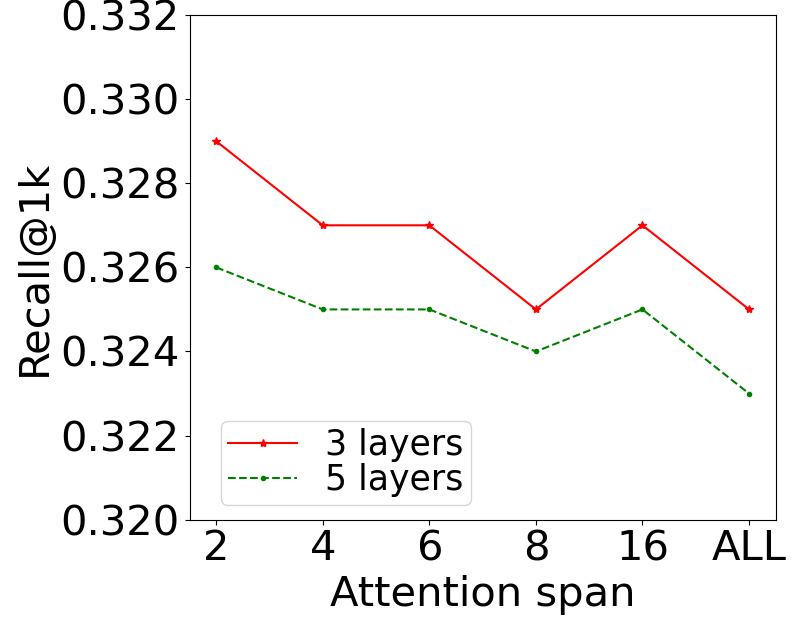}
\end{minipage}
}
\subfigure[Recall@1k\label{fig:recall}]{
\begin{minipage}[t]{0.465\linewidth}
\centering
\includegraphics[width=3.8cm]{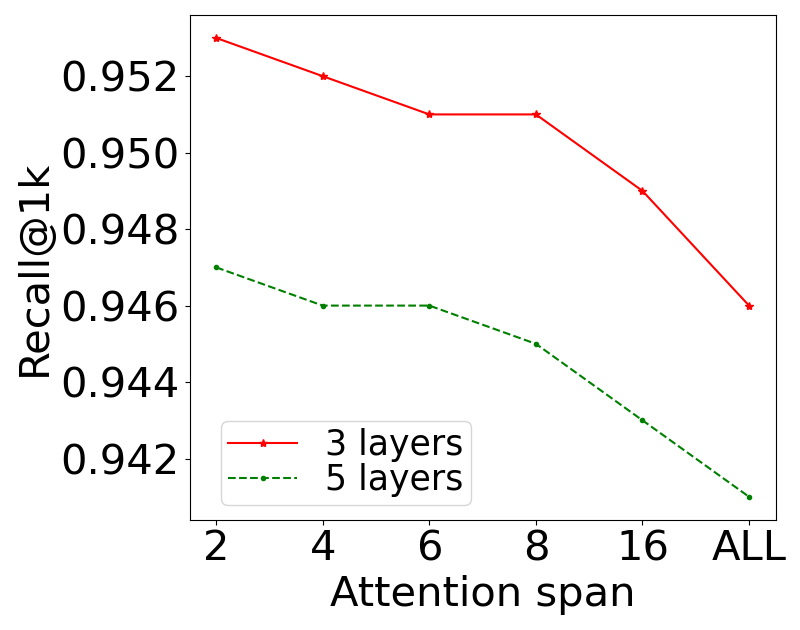}
\end{minipage}
}
\caption{MS MARCO passage Dev accuracy of Siamese (BM25 Neg) when fine-tuned from \model variations.
\label{fig:ablation}}
\vspace{-0.4cm}
\end{figure}

%% file: Figures/dependency.tex
\begin{figure}[t]
\centering
\subfigure[Full Attention.\label{fig:dependency}]{
\begin{minipage}[t]{0.465\linewidth}
\centering
\includegraphics[width=3.8cm]{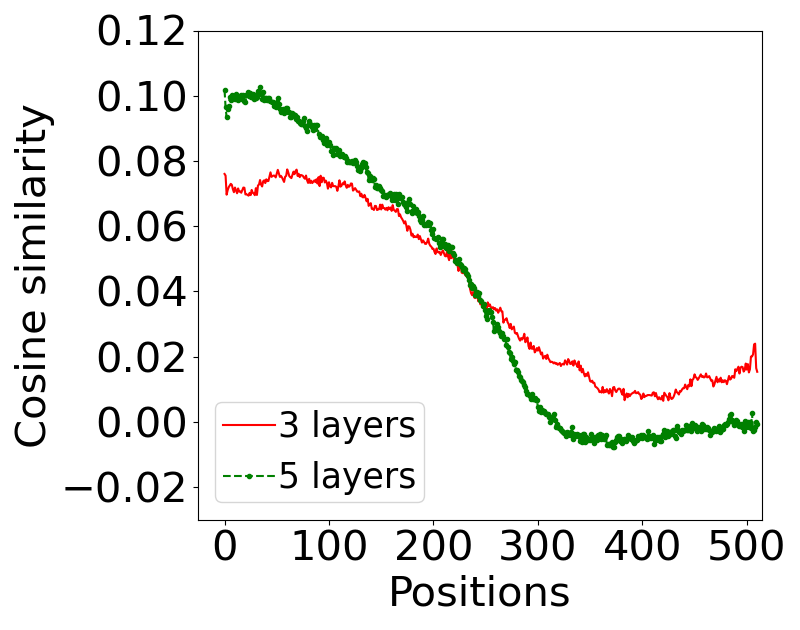}
\end{minipage}
}
\subfigure[Restricted Attention.\label{fig:dependency_window}]{
\begin{minipage}[t]{0.465\linewidth}
\centering
\includegraphics[width=3.8cm]{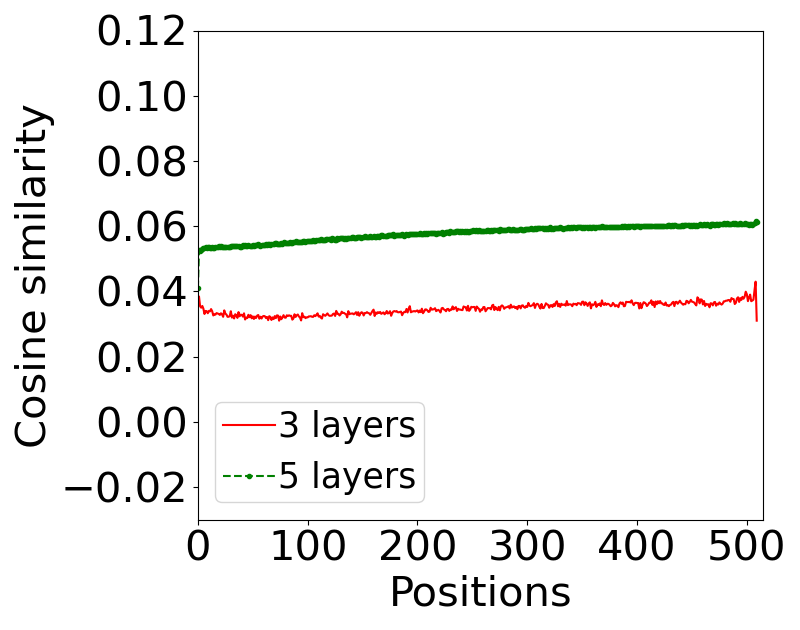}
\end{minipage}
}
\caption{The cosine similarity between encoder \texttt{CLS} and the token representations from the decoder at different positions: 0 is the beginning of the sequence and the closest to \texttt{CLS}. The restricted attention sets attention span to two.
} 
\label{fig:whole_dependency}
\vspace{-0.3cm}
\end{figure}

%% file: Figures/compare_optimus.tex
\begin{figure}[t]
\centering
\includegraphics[width=3.8cm,center]{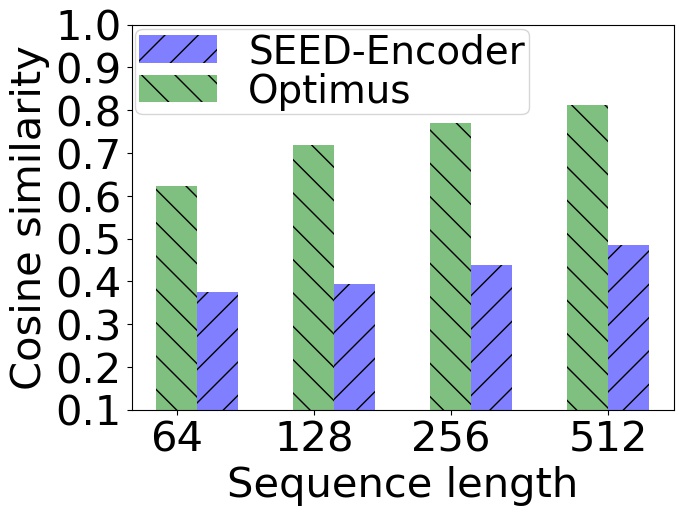}
\caption{Cosine similarity of sequences with different lengths using Optimus and \model.
\label{fig:compare_optimus}}

\vspace{-0.4cm}
\end{figure}

%% file: Figures/startingpoint.tex
\begin{figure*}[t]
\centering
\subfigure[Siamese (BM25 Neg).\label{fig:s_bm25}]{
\begin{minipage}[t]{0.23\linewidth}
\centering
\includegraphics[width=3.87cm]{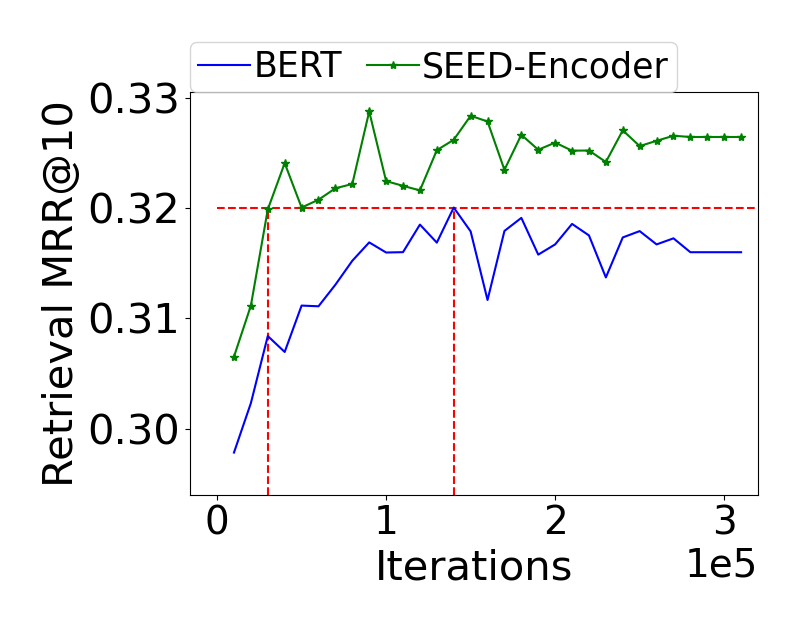}
\end{minipage}
}
\subfigure[ANCE (FirstP).\label{fig:s_ance}]{
\begin{minipage}[t]{0.23\linewidth}
\centering
\includegraphics[width=3.87cm]{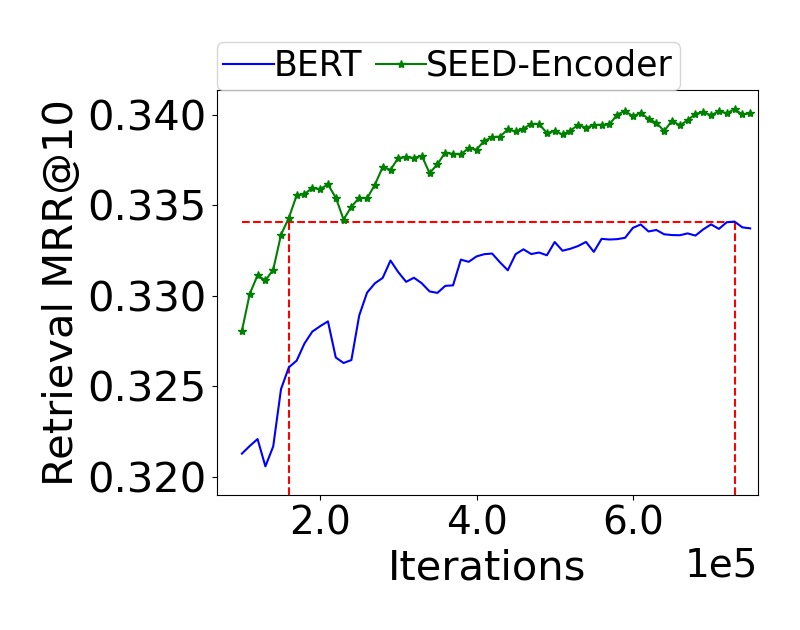}
\end{minipage}
}
\subfigure[Siamese (BM25 Neg).\label{fig:fewshot_bm25}]{
\begin{minipage}[t]{0.23\linewidth}
\centering
\includegraphics[width=3.87cm]{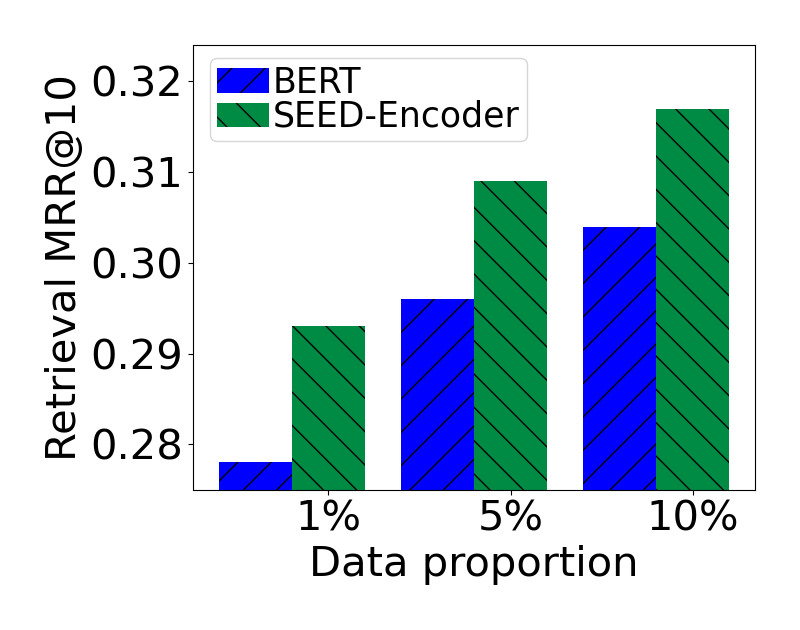}
\end{minipage}
}
\subfigure[ANCE (FirstP).\label{fig:fewshot_ance}]{
\begin{minipage}[t]{0.23\linewidth}
\centering
\includegraphics[width=3.87cm]{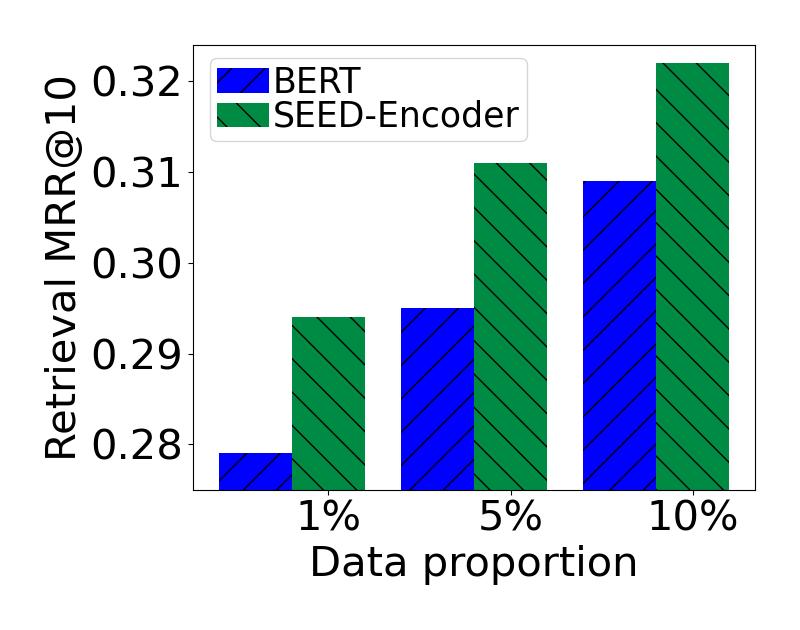}
\end{minipage}
}
\caption{MS MARCO passage retrieval accuracy of Siamese (BM25 Neg) and ANCE (FirstP) when fine-tuned from BERT (Ours) and \model. (a) and (b) are their accuracy at different fine-tuning steps (x-axes, in 100K). (c) and (d) are their accuracy with a fraction (x-axes) of training labels in the few-shot setting.}
\label{fig:retrieval_result}
\vspace{-0.3cm}
\end{figure*}

%% file: tables/case_study.tex
\begin{table*}[t]
\vspace{0.3cm}
\small
\centering
\resizebox{1.0\textwidth}{!}{
\begin{tabular}{ll|l}
\hline
    {} & \textbf{Case 1} & \textbf{Case 2}\\
    \hline
      \textbf{Query} &
     hiking on mount rainier in the winter & \makecell*[l]{what kind of party is the cooperative party}  \\
    \hline
    SEED-Encoder & MRR@100 1.0 &  MRR@100 1.0 \\
     \hline
    \textbf{Url} & \makecell*[l]{ https://www.nps.gov/mora/planyourvisit/winter-recreation.htm  } & \makecell*[l]{ https://simple.wikipedia.org/wiki/Co-operative\_Party}\\
    \hline
    \textbf{Title} & Winter Recreation & Cooperative Party  \\
    \hline
    \textbf{Snippet} & \makecell*[l]{Winter Recreation Winter Camping Food Storage Snowplay... \\\textbf{A Winter visit to Mount Rainier} can include ranger-guided \\snowshoe walks, skiing...Learn about \textbf{winter hiking} opportu-\\nities at Longmire in...} & \makecell*[l]{\textbf{Co-operative Party} From Wikipedia, the free encyclopedia \\navigation search. The \textbf{Co-operative Party} is a small socia-\\list political party, in the United Kingdom. Its candidates \\must be members of the Labour Party as well...} \\
    \hline
    RoBERTa & MRR@100 0.043 & MRR@100 0.067 \\
     \hline
    \textbf{Url} & \makecell*[l]{ http://www.seattletimes.com/life/travel/5-great-day-hikes-\\around-mount-rainier/} & \makecell*[l]{ http://socialeconomyaz.org/whats-a-cooperative/}\\
    \hline
    \textbf{Title} & 5 great day-hikes around Mount Rainier  & What is a Cooperative? \\
    \hline
    \textbf{Snippet} & \makecell*[l]{Life Outdoors Travel5 great day-hikes around Mount Rainier\\ Originally published June 24, 2015 at 4:59...(Picasa)E-book \\authors name their favorite day-hikes in Mount Rainier Na-\\tional Park...  } & \makecell*[l]{What is a Cooperative? According to the International Coo-\\perative Alliance ( ICA ), a cooperative is "an autonomous \\association of persons united voluntarily to meet their com-\\mon economic, social, and cultural needs...}\\
    \hline
 \end{tabular}
 }
\caption{Two examples of \model's winning case over RoBERTa (Ours) when fine-tuning with ANCE FirstP in MARCO Document. Their first ranked documents are listed.}
\label{tab:case}
\end{table*}

%% file: 05_conclusion.tex
\section{Conclusion}

In this paper we present \model, a self-training framework dedicated  to pre-training language models for dense text retrieval. We pre-train an auto-encoder that employs a weak decoder with restricted capacity and attention span following our mathematical derivation. The weak decoder helps \model capture more context information and generate better text representation. In our experiments on web search, news recommendation, and question answering, \model initialized dense retrieval models achieve state-of-the-art accuracy compared to several strong baselines. Future work along this direction includes exploring more self-learning tasks and network architectures for sequence matching in dense retrieval scenarios.



%% file: 07_acknowledgements.tex
\section*{Acknowledgements}

We would like to thank anonymous reviewers for
their valuable comments. This work is partially supported by National Natural Science Foundation of China NO. 61872370,  and Beijing Outstanding Young Scientist Program NO. BJJWZYJH012019100020098.

%% file: 06_appendix.tex
\section{Appendix}





\input{tables/MSMARCO}
\input{tables/MIND}

\subsection{More Details of MS MARCO, MIND and OpenQA dataset}
\label{app:dataset}
\paragraph{More Details of MARCO Dataset}

Microsoft MARCO~\citep{bajaj2016ms} is the largest available search benchmark to date. It includes two tasks, document ranking and passage ranking. Both are to find and rank relevant documents/passages from a web corpus for a web query from Bing. The dataset statistics are summarized in Table~\ref{tab:MSMARCO}.

\paragraph{More Details of MIND Dataset}
MIcrosoft News Dataset (MIND)~\cite{wu2020mind} is a large-scale recommendation dataset that collects about 160k English news articles and more than 15 million user impression logs from MSN news. Each news article contains the title, abstract, body, and category. Each impression log includes the user's click behavior on the page and her historical news click behaviors. The task is to rank a given set of candidate news articles, e.g., those from an early stage of their recommendation pipeline, based on the user's previous click history. The dataset statistics are summarized in Table~\ref{tab:MIND}.

\paragraph{More Details of NQ Dataset}
For OpenQA experiments we use the Natural Question query set~\cite{kwiatkowski2019natural}, in which the queries are mined from real
Google search queries and the corresponding answers are spans
in Wikipedia articles identified by annotators. We use the Wikipedia passages preprocessed and shared in  DPR~\cite{karpukhin2020dense}, which includes $21,015,324$ passages. More detailed data such as the number of queries can be found in \citet{karpukhin2020dense}

\input{tables/glue}
\subsection{GLUE}
\label{sec:glue}
We also consider the GLUE benchmark~\cite{wang2018glue} which contains nine datasets for general language understanding. Here we select MNLI, QQP, QNLI and SST-2 from the GLUE benchmark, and compare the performance of \model with BERT (Ours) and Optimus on these tasks. We follow the fine-tuning schedule in \citet{devlin2018bert}, and the results are shown in Table~\ref{tab:glue}. We can see that on these GLUE tasks, \model is not worse than BERT and Optimus. This shows that while \model can generate higher-quality representations that well fit the Siamese network, the performance on GLUE will not become worse.

%% file: tables/MSMARCO.tex
\begin{table}[t]
\vspace{-0.1cm}
\small
\centering
\resizebox{0.48\textwidth}{!}{
\begin{tabular}{l|rrr|rrr}
\hline
    {} & \multicolumn{3}{c|}{\textbf{Document}} & \multicolumn{3}{c}{\textbf{Passage}}\\
    \hline
    \textbf{Model} &
    \multicolumn{1}{c|}{\textbf{Train}} &
    \multicolumn{1}{c|}{\textbf{Dev}} &
     \multicolumn{1}{c|}{\textbf{Eval}} & \multicolumn{1}{c|}{\textbf{Train}} & \multicolumn{1}{c|}{\textbf{Dev}} &
      \multicolumn{1}{c}{\textbf{Eval}} \\
    \hline
    Query & 367,013	 & 5,193  & 5,793 & 808,731 & 101,093 & 101,092\\
    relevant label & 384,597 & 5,478 & - & 532,761  & 59,273	 & - \\ 
    Doc set & \multicolumn{3}{c|}{3,213,835	} & \multicolumn{3}{c}{8,841,823}\\
    \hline
\end{tabular}
}
\caption{Statistics of the MSMARCO dataset
}
\label{tab:MSMARCO}
\end{table}

%% file: tables/MIND.tex
\begin{table}[t]
\small
\centering
\resizebox{0.4\textwidth}{!}{
\begin{tabular}{l|rr}
\hline
    \textbf{} & \textbf{Train} & \textbf{Dev}\\
    \hline
    Users & 711,222 & 255,990\\
    News & 101,527 & 72,023\\
    Impression & 2,232,748 & 376,471 \\
    Avg. title len &14.41 &14.47\\
    Avg. click num & 1.52 & 1.53 \\
    Avg. candidate num& 37.40 & 37.41\\
    Avg. historical news click num &32.98 &32.62\\
    \hline
 \end{tabular}
 }
\caption{Statistics of the MIND dataset
}
\label{tab:MIND}
\end{table}

%% file: tables/glue.tex
\begin{table}[t]
\vspace{-0.1cm}
\small
\centering
\resizebox{0.45\textwidth}{!}{
\begin{tabular}{l|r|r|r|r}
\hline
    model &  MNLI & QQP & SST-2 & QNLI  \\
    \hline
    BERT (Ours) & 0.849 & 0.910 & 0.929 & 0.913 \\
    Optimus & 0.834 & 0.909 & 0.923 & 0.912 \\
    \hline
    SEED-Encoder & 0.843 & 0.911 & 0.927 & 0.914\\
    \hline
\end{tabular}
}
\caption{Results on some GLUE tasks.
}
\label{tab:glue}
\vspace{-0.3cm}
\end{table}